# CenterFace: Joint Face Detection and Alignment Using Face as Point

Yuanyuan Xu[1,4]    Wan Yan[2]    Haixin Sun[3]    Genke Yang[4]    Jiliang Luo[1]

[1]Huaqiao University    [2]StarClouds    [3]Xiamen University    [4]Shanghai Jiaotong University

**Abstract**. Face detection and alignment in unconstrained environment is always deployed on edge devices which have limited memory storage and low computing power. This paper proposes a one-stage method named CenterFace to simultaneously predict facial box and landmark location with real-time speed and high accuracy. The proposed method also belongs to the anchor free category. This is achieved by: (a) learning face existing possibility by the semantic maps, (b) learning bounding box, offsets and five landmarks for each position that potentially contains a face. Specifically, the method can run in real-time on a single CPU core and 200 FPS using NVIDIA 2080TI for VGA-resolution images, and can simultaneously achieve superior accuracy (WIDER FACE Val/Test-Easy: 0.935/0.932, Medium: 0.924/0.921, Hard: 0.875/0.873 and FDDB discontinuous: 0.980, continuous: 0.732). A demo of CenterFace can be available at https://github.com/Star-Clouds/CenterFace.

## 1 Introduction

Face detection and alignment is one of the fundamental issues in computer vision and pattern recognition, and is often deployed in mobile and embedded devices. These devices typically have limited memory storage and low computing power. Therefore, it is necessary to predict the position of the face box and the landmark at the same time, and it is excellent in speed and precision.

With the great breakthrough of convolutional neural networks(CNN), face detection has recently achieved remarkable progress in recent years. Previous face detection methods have inherited the paradigm of anchor-based generic object detection frameworks, which can be divided into two categories: two-stage method (Faster-RCNN [15]) and one-stage method (SSD [17]). Compared with two-stage method, the one-stage method is are more efficient and has higher recall rate, but it tends to achieve a higher false positive rate and to compromise the localization accuracy. Then Hu et al. [7] used a two-stage approach to the Region Proposal Networks (RPN) [15] to detect faces directly, while SSH [8] and S3FD [10] developed a scale-invariant network in a single network to detect faces with mutil-scale from different layers.

The previous anchor-based methods have some drawbacks. On the one hand, in order to improve the overlap between anchor boxes and ground truth, a face detector usually requires a large number of dense anchors to achieve a good recall rate. For example, more than 100k anchor boxes in RetinaFace [11] for a $640\times640$ input image. On the other hand, the anchor is a hyperparameter design that is statistically calculated from a particular data set, so it is not always feasible to other applications, which goes against the generality.

* E-Mail: hixyy@126.com, yanwan@starcloudsglobal.com.cn

In addition, the current state of the art FACE detectors has achieved considerable accuracy on the benchmark WIDER FACE [20] by using heavy pretrained backbones such as VGG16 [21] and resnet50/152 [22]. First, these detectors are difficult to use in practice because the network consumes too much time and the model size is also too large. Secondly, it is not convenient for face recognition application without facial landmark prediction. Therefore, joint detection and alignment, as well as better balance of accuracy and latency, are essential for practical applications.

Inspired by the anchor free universal object detection framework [1, 3, 6, 14, 15, 25, 26], this paper proposes a simpler and more effective face detection and alignment method named CenterFace, which is only lightweight but also powerful. The network structure about the CenterFace is shown in Figure 1, which can be trained end-to-end. We use the center point of the face's bounding box to represent the face, then facial box size and landmark are regressed directly to image features at the center location. So face detection and alignment are transformed to the standard key point estimation problem [4, 27, 28]. The peak in the heat map corresponds to the center of the face. The image features at each peak predict the size of the face and the face key points. This approach was fully evaluated and the latest detection performance were shown on a number of benchmark data sets for face detection, including FDDB [24] and WIDER FACE.

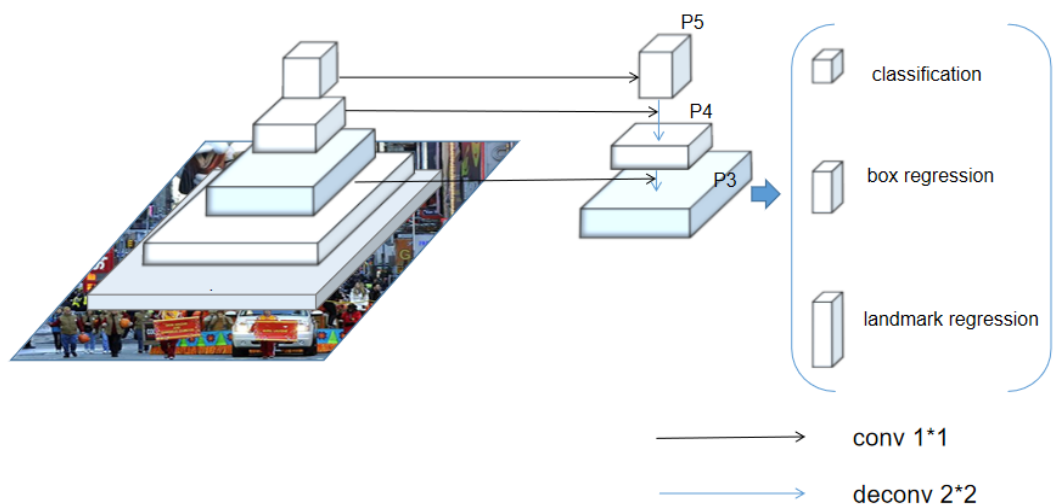

Figure 1 Architecture of the CenterFace

In summary, the main contributions of this work can be summarized as four-fold:

- By introducing the anchor free design, face detection is transformed into a standard key point estimation problem, using only a larger output resolution (output stride is 4) compared to previous detectors.

- Based on the multi-task learning strategy, the face as point design is proposed to predict the face boxes and five key points at the same time.

- This paper proposes a feature pyramid network using common Layer for accurate and fast face detection.

- Comprehensive experimental results based on popular benchmarks FDDB and WIDER FACE, as well as CPU and GPU hardware platforms, have demonstrated the superiority of the proposed method in terms of speed and accuracy.

## 2 Related Works

### 2.1 Cascaded CNN methods

The method of cascade convolutional neural network (CNN) [29, 30, 31] uses cascaded CNN framework to learn features in order to improve the performance and maintain efficiency. However, there are some problems about cascaded CNN based detector: 1) The runtime of these detector is negatively correlated with the number of faces on the input image. The speed will dramatically degrade when the number of faces increases; 2) Because these methods optimize each module separately, the training process becomes extremely complicated.

### 2.2 Anchor methods

Inspired by generic object detection methods [16, 17, 25, 26, 35, 36, 37, 38], which embraced all the recent advancement in deep learning, face detection has recently achieved remarkable progress [7, 8, 9, 10]. Different from generic object detection, the ratio of face scale is usually from 1:1 to 1:1.5. The latest methods [9, 11] focus on single stage design, which densely samples face locations and scales on feature pyramids, demonstrating promising performance and yielding faster speed compared to two-stage methods [18, 19].

### 2.3 Anchor free methods

In our view, Cascaded CNN methods are also a kind of anchor free methods. However, this method uses sliding window to detect human faces and relies on image pyramids. It has some shortcomings such as slow speed and complex training process. LFFD[12] regards the RFs as natural anchors which can cover continuous face scales, which is just another way to define anchor, but the training time is about 5 days with two NVIDIA GTX1080TI. Our CenterFace simply represents faces by a single point at their bounding box center, then facial box size and landmark are regressed directly from image features at the center location. Thus face detection is transformed into a standard key point estimation problem. And the training time of a NVIDIA GTX2080TI is only one day.

### 2.4 Multitask Learning

Multitask learning uses multiple supervisory labels to improve the accuracy of each task by utilizing the correlation between tasks. Joint face detection and alignment [27, 29] is widely used because alignment task，paralleling with the backbone, provides better features for face classification task with face point information. Similarly, Mask R-CNN [5] significantly improves the detection performance by adding a branch for predicting an object mask.

## 3 CenterFace

### 3.1 Mobile Feature Pyramid Network

We adopted Mobilenetv2 [32] as the backbone and Feature Pyramid Network (FPN) [25] as the neck for the subsequent detection. In general, FPN uses a top-down architecture with lateral connections to build a feature pyramid from a single scale input. CenterFace represents the face through the center point of face box, and face size and facial landmark are then regressed directly from image features of the center location. Therefore only one layer in the pyramid is used for face detection and alignment. We construct a pyramid with levels {P-L}, L = 3, 4, 5, where L indicates pyramid level. Pl has $1/2^L$ resolution of the input. All pyramid levels have C = 24 channels.

## 3.2 Face as Point

Let $[x_1, y_1, x_2, y_2]$ be the bounding box of face. Facial center point lies at $^c = [(x_1+ x_2)/2, (y_1+ y_2)/2]$. Let $I \in R^{W \times H \times 3}$ be an input image of width W and height H. Our aim is to produce the heatmap $Y \in [0, 1]^{W/R \times H/R}$, where R is the output stride. We use the default output stride of R = 4 in literature [23]. A prediction $\hat{Y}_{x,y} = 1$ corresponds to a face center, while $\hat{Y}_{x,y} = 0$ is background.

The face classification branch is trained following Law and Deng [23]. For each ground truth. We calculate the equivalent heat map by using Gaussian kernel to represent the ground truth. The training loss is a variant of focal loss [26]:

$$L_c = \begin{cases} -(1-\hat{Y}_{xyc})^\alpha \log(\hat{Y}_{xyc}) & if\ Y_{xyc} = 1 \\ -(1-Y_{xyc})^\beta (\hat{Y}_{xyc})^\alpha \log(1-\hat{Y}_{xyc}) & otherwise \end{cases} \quad (1)$$

where $\alpha$ and $\beta$ are hyper-parameters of the focal loss, which are designated as $\alpha = 2$ and $\beta = 4$ in all our experiments.

To gather global information and to reduce memory usage, downsampling is applied to an image convolutionally, the size of the output is usually smaller than the image. Hence, a location $(x, y)$ in the image is mapped to the location $(x/n, y/n)$ in the heatmaps, where $n$ is the downsampling factor. When we remap the locations from the heatmaps to the input image, some pixel may be not alignment, which can greatly affect the accuracy of facial boxes. To address this issue, we predict position offsets to adjust the center position slightly before remapping the center position to the input resolution：

$$o_k = (\frac{x_k}{n} - \lfloor \frac{x_k}{n} \rfloor, \frac{y_k}{n} - \lfloor \frac{y_k}{n} \rfloor) \quad (2)$$

where $o_k$ is the offset, $x_k$ and $y_k$ are the $x$ and $y$ coordinate for face center $k$. We apply the L1 Loss [5] at ground-truth center position:

$$L_{off} = \frac{1}{N} \sum_{k=1}^{N} SmoothL1\_Loss(o_k, \hat{o}_k) \quad (3)$$

## 3.3 Box and Landmark Prediction

To reduce the computational burden, we use a single size prediction $S \in R^{W/4*H/4}$ for facial box and landmarks. Each ground-truth bounding box is specified as $G = (x1, y1, x2, y2)$. Our goal is to learn a transformation that maps the networks position outputs $(\hat{h}, \hat{w})$ to center position $(x, y)$ in the feature maps:

$$\hat{h} = \log(\frac{x_2}{R} - \frac{x_1}{R})$$
$$\hat{w} = \log(\frac{y_2}{R} - \frac{y_1}{R}) \quad (4)$$

Different from Box regression, the regression of the five facial landmarks adopts the target normalization method based on the center position:

$$lm_{\hat{x}} = \frac{lm_x}{box_w} - \frac{c_x}{box_w}$$
$$lm_{\hat{y}} = \frac{lm_y}{box_h} - \frac{c_y}{box_h} \quad (5)$$

We also use smooth L1 loss to facial box and landmarks prediction at the center location

$$L = L_c + \lambda_{off} L_{off} + \lambda_{box} L_{box} + \lambda_{lm} L_{lm} \quad (6)$$

Where $\lambda_{off}$, $\lambda_{box.}$ and $\lambda_{lm}$ is used to scale the loss .We use *1, 0.1, 0.1*, respectively in all our experiments.

### 3.4. Training Details

**Dataset**. The proposed method is trained on the training set of WIDER FACE benchmark, including 12,880 images with more than 150,000 valid faces in scale, pose, expression, occlusion and illumination. RetinaFace [11] introduces five levels of face image quality and annotates five landmarks on faces.

**Data augmentation.** Data augmentation is important to improve the generalization. We use random flip, random scaling [33], color jittering and randomly crop square patches from the original images and resize these patches into 800*800 to generate larger training faces. Faces that are less than 8 pixels are discarded directly.

**Training parameters**. We train the CenterFace using Adam optimiser with a batch-size 8 and learning rate 5e-4 for 140 epochs, with learning rate dropped 10× at 90 and 120 epochs, respectively. The down-sampling layers of MobilenetV2 are initialized with ImageNet pretrain and the up-sampling layers are randomly initialized. The training time is about one day with one NVIDIA GTX2080TI.

## 4 Experiments

In this section, we firstly introduce the runtime efficiency of CenterFace, then evaluate it on the common face detection benchmarks.

### 4.1 Running Efficiency

The existing CNN face detectors can be accelerated by GPUs, but they are not fast enough in most practical applications, especially CPU based applications. As described below, our CenterFace is efficient enough to meet practical requirements and its model size is only 7.2MB. In Table 1, comparing with other detectors, our method can exceed the real-time running speed (> 100 FPS) at different resolutions by using a single NVIDIA GTX2080TI.

Owing to the DSFD, PyramidBox, S3FD and SSH are too slow when running on CPU platforms, we only evaluate the proposed CenterFace, FaceBoxes, MTCNN and CasCNN at VGA-resolution images on CPU and the mAP means the true positive rate at 1000 false positives on FDDB. As listed in Table 2, our CenterFace can run at 30 FPS on the CPU with state-of-the-art accuracy.

Table 1. Running efficiency on GTX2080TI

| Approach | **640*480** | 1280*720 | 1920*1080 |
|---|---|---|---|
| DSFD | 78.08ms | 187.78ms | 393.82ms |
| PyramidBox | 50.51ms | 142.34ms | 331.93ms |
| S3FD | 21.75ms | 55.73ms | 119.53ms |
| LFFD | 7.60ms | 16.37ms | 31.41ms |
| CenterFace | 5.51ms | 6.47ms | 8.79 ms |

Table 2. Running efficiency on CPU

| Approach | **CPU-model** | mAP(%) | FPS |
|---|---|---|---|
| CasCNN | E5-2620@2.00 | 85.7 | 14 |
| MTCNN | N/A@2.60 | 94.4 | 16 |
| Faceboxes3.2 | E5-2660v3@2.60 | 96.0 | 20 |
| CenterFace | I7-6700@2.6 | 98.0 | 30 |

## 4.2 Evaluation on Benchmarks

**FDDB dataset**. FDDB contains 2845 images with 5171 unconstrained faces collected from the Yahoo news website. We evaluate our face detector on FDDB against the other state-of-art methods and the results are shown in Table 3 and Fig. 2, respectively. We also add DFSD, PyramidBox and S3FD detectors. Whereas, these detectors are much slower due to the larger backbone and denser anchors. Our CenterFace can also achieve good performance on both discontinuous and continuous ROC curves, i.e. 98.0% and 72.9% when the number of false positives equals to 1,000 and it outperforms LFFD, FaceBoxes and MTCNN evidently.

Table 3. Evaluation results on FDDB

| Method | Disc ROC curves score | Cont ROC curves score |
|---|---|---|
| DFSD | 0.984 | 0.754 |
| PyramidBox | 0.982 | 0.757 |
| S3FD | 0.981 | 0.754 |
| MTCNN | 0.944 | 0.708 |
| Faceboxes | 0.960 | 0.729 |
| LFFD | 0.973 | 0.724 |
| CenterFace | 0.980 | 0.732 |

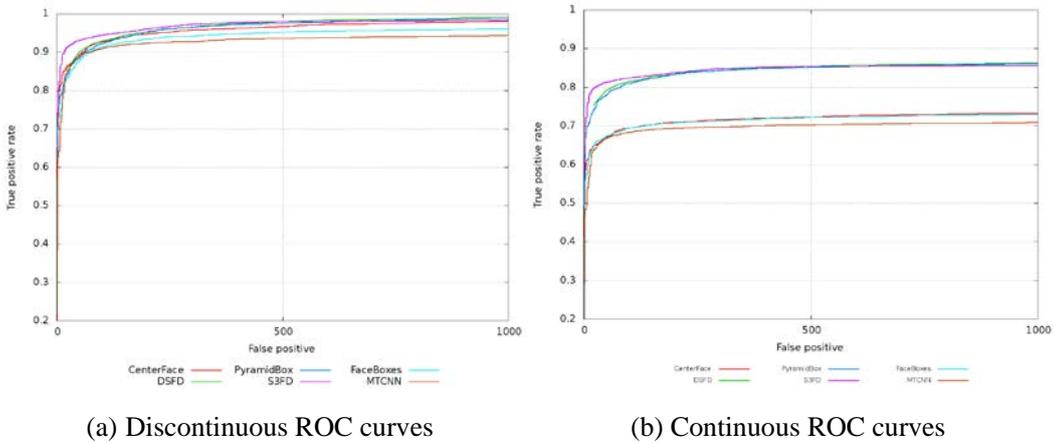

(a) Discontinuous ROC curves        (b) Continuous ROC curves

Figure 2. Evaluation on the FDDB dataset.

**WIDER FACE dataset**. Until now, WIDER FACE is the most widely used benchmark for face detection. The WIDER FACE dataset is split into training (40%), validation (10%) and testing (50%) subsets by randomly sampling from 61 scene categories. All the compared methods are trained on training set. For testing on WIDER FACE, we follow the standard practices of [11] and employ flip as well as multi-scale strategies. Box voting [13] is applied on the union set of predicted face boxes using an IoU threshold at 0.4. We report the results on the validation and testing sets in Tables 4 and 5, respectively. The proposed method CenterFace achieves 0.935 (Easy), 0.924 (Medium) and 0.875 (Hard) for validation set, and 0.932 (Easy), 0.921 (Medium) and 0.873 (Hard) for testing set. Although it has gaps with state of the art methods, but consistently outperforms SSH (using VGG16 as the backbone), LFFD, FaceBoxes and MTCNN. Additionally, CenterFace is better than S3FD that uses VGG16 as the backbone and dense anchors on Hard parts.

Furthermore, we also test on WIDER FACE not only with the original image but also with a single inference, our CenterFace also produces the good average precision (AP) in all the subsets of both validation sets, i.e., 92.2% (Easy), 91.1% (Medium) and 78.2% (Hard) for validation set.

Table 4. Performance results on the validation set of WIDER FACE.

| Method | Easy | Medium | Hard |
|---|---|---|---|
| RetinaFace | 0.969 | 0.961 | 0.918 |
| DSFD | 0.966 | 0.957 | 0.904 |
| PramidBox | 0.961 | 0.950 | 0.889 |
| S3FD | 0.937 | 0.924 | 0.852 |
| SSH | 0.931 | 0.921 | 0.845 |
| MTCNN | 0.848 | 0.825 | 0.598 |
| Faceboxes | 0.840 | 0.766 | 0.395 |
| LFFD | 0.910 | 0.881 | 0.780 |
| CenterFace | 0.935 | 0.924 | 0.875 |

Table 5. Performance results on the testing set of WIDER FACE.

| Method | Easy | Medium | Hard |
|---|---|---|---|
| RetinaFace | 0.963 | 0.956 | 0.914 |
| DSFD | 0.960 | 0.953 | 0.900 |
| PramidBox | 0.956 | 0.946 | 0.887 |
| S3FD | 0.928 | 0.913 | 0.840 |
| SSH | 0.927 | 0.915 | 0.844 |
| MTCNN | 0.851 | 0.820 | 0.607 |
| Faceboxes | 0.839 | 0.763 | 0.396 |
| LFFD | 0.896 | 0.865 | 0.770 |
| CenterFace | 0.932 | 0.921 | 0.873 |

## 5 Conclusion

This paper introduces the CenterFace that has the superiority of the proposed method perform well on both speed and accuracy and simultaneously predicts facial box and landmark location. Our proposed method overcomes the drawbacks of the previous anchor based method by translating face detection and alignment into a standard key point estimation problem. CenterFace represents the face through the center point of face box, and face size and facial landmark are then regressed directly from image features of the center location. Comprehensive and extensive experiments are made to fully analyze the proposed method. The final results demonstrate that our method can achieve real-time speed and high accuracy with a smaller model size, making it an ideal alternative for most face detection and alignment applications.

**Acknowledgments** This work was supported in part by the National Key R&D Program of China (2018YFC0809200) and the Natural Science Foundation of Shanghai (16ZR1416500).

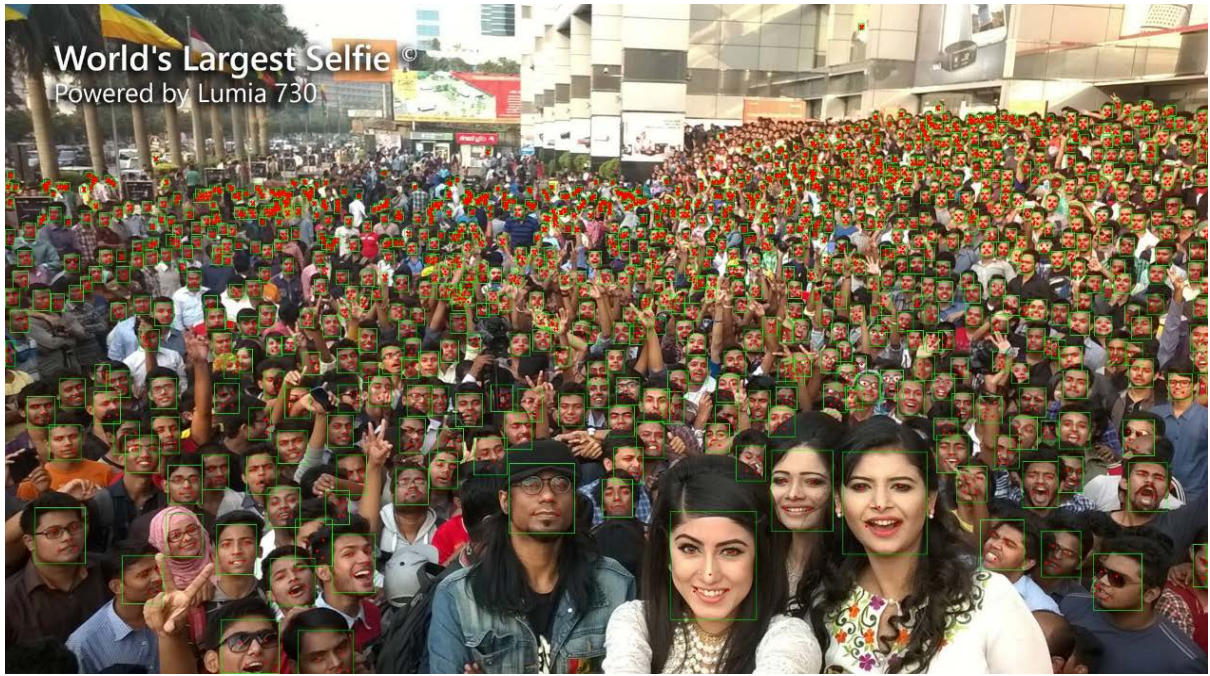

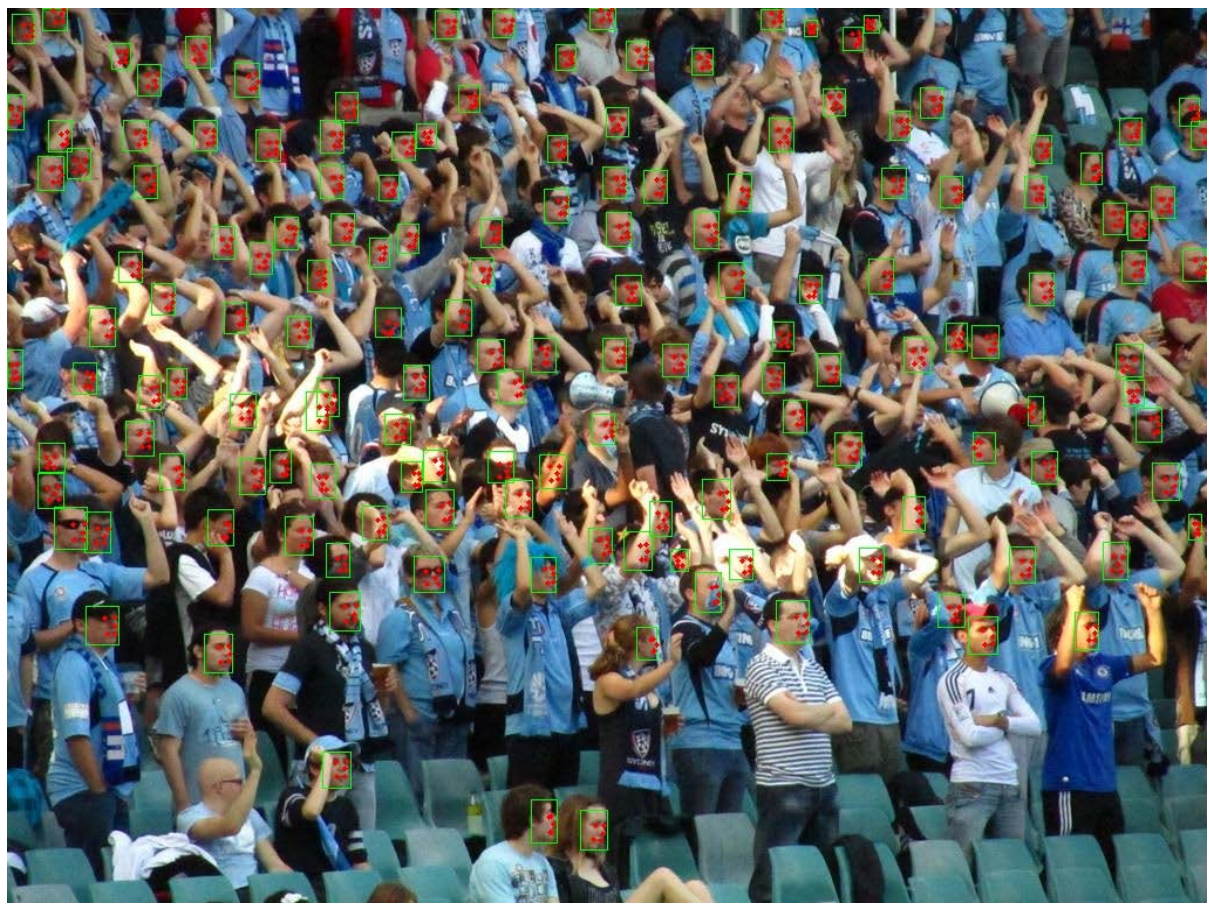

Figure 3. Face Detection Results on WIDER FACE.